\documentclass{article}

\usepackage[nonatbib,preprint]{neurips_2024}

\usepackage[utf8]{inputenc} % allow utf-8 input
\usepackage[T1]{fontenc}    % use 8-bit T1 fonts
\usepackage{hyperref}       % hyperlinks
\usepackage{url}            % simple URL typesetting
\usepackage{booktabs}       % professional-quality tables
\usepackage{amsfonts}       % blackboard math symbols
\usepackage{nicefrac}       % compact symbols for 1/2, etc.
\usepackage{microtype}      % microtypography
\usepackage{xcolor}         % colors

\usepackage{hyperref}
\usepackage{graphicx}
\usepackage{rotating}

\usepackage{orcidlink}
\usepackage{multirow}
\usepackage{mathrsfs}
\usepackage{amsmath,amssymb}
\usepackage{makecell}
\usepackage[ruled,noend,linesnumbered]{algorithm2e}

\usepackage{enumitem}

\usepackage{wrapfig}

\usepackage{caption}
\usepackage{subcaption}

\usepackage{floatrow}
\newfloatcommand{capbtabbox}{table}[][\FBwidth]

\usepackage[
backend=biber,
sorting=anyt,
bibstyle=ieee,
citestyle=numeric,
]{biblatex}
\title{Dynamic Object Queries for Transformer-based Incremental Object Detection}

\author{%
  Jichuan Zhang$^{1}$\footnotemark[1],  Wei Li$^{1}$\footnotemark[1],  Shuang Cheng$^{2}$,  Ya-Li Li$^{1}$\footnotemark[2], Shengjin Wang$^{1}$\\
  $^1$Department of Electronic Engineering, Tsinghua University\\
  $^2$Institute of Computing Technology, Chinese Academy of Sciences, Beijing, China\\
  University of Chinese Academy of Science\\
  \texttt{\{zhangjc22, lw22\}@mails.tsinghua.edu.cn}\\
  \texttt{chengshuang22@mails.ucas.ac.cn}\\
  \texttt{\{liyali13, wgsgj\}@tsinghua.edu.cn}\\ 
}

\begin{document}

\maketitle
\renewcommand{\thefootnote}{\fnsymbol{footnote}} %将脚注符号设置为fnsymbol类型，即特殊符号表示
\footnotetext[1]{These authors contributed equally to this work.} %对应脚注[1]
\footnotetext[2]{Corresponding authors.} %对应脚注[2]

\begin{abstract}
Incremental object detection (IOD) aims to sequentially learn new classes, while maintaining the capability to locate and identify old ones. As the training data arrives with annotations only with new classes, IOD suffers from catastrophic forgetting. Prior methodologies mainly tackle the forgetting issue through knowledge distillation and exemplar replay, ignoring the conflict between limited model capacity and increasing knowledge. In this paper, we explore \textit{dynamic object queries} for incremental object detection built on Transformer architecture. We propose the \textbf{Dy}namic object \textbf{Q}uery-based \textbf{DE}tection \textbf{TR}ansformer (DyQ-DETR), which incrementally expands the model representation ability to achieve stability-plasticity tradeoff. First, a new set of learnable object queries are fed into the decoder to represent new classes. These new object queries are aggregated with those from previous phases to adapt both old and new knowledge well. Second, we propose the isolated bipartite matching for object queries in different phases, based on disentangled self-attention. The interaction among the object queries at different phases is eliminated to reduce inter-class confusion. Thanks to the separate supervision and computation over object queries, we further present the risk-balanced partial calibration for effective exemplar replay. Extensive experiments demonstrate that DyQ-DETR significantly surpasses the state-of-the-art methods, with limited parameter overhead. Code will be made publicly available.
\end{abstract}

\section{Introduction}
\label{sec:intro}

\begin{figure}[tb]
  \centering
  \includegraphics[height=5.0cm]{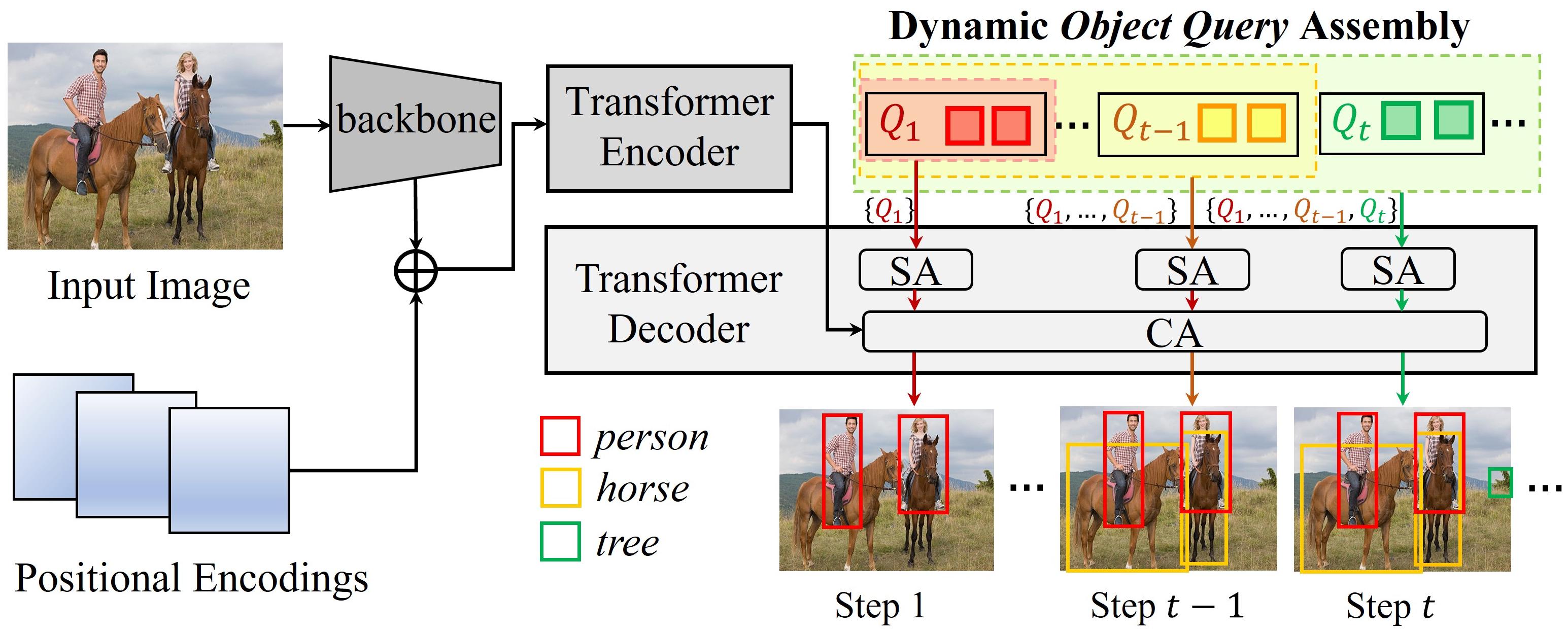}
  \caption{\textbf{Illustration of DyQ-DETR.} Built on Detection Transformer, a new set of queries is assigned for the newly-arriving classes at each step and different groups of queries are responsible for detecting specific classes annotated in corresponding steps. $Q_t$ denotes the query group in step $t$. SA and CA refer to the self-attention and cross-attention modules, respectively.}
  \label{fig:intro}
\end{figure}

Humans inherently possess the ability to incrementally learn novel concepts without forgetting previous ones, capable of acquiring and accumulating knowledge from past experiences. Traditional object detection models \cite{girshick2015fast,zhou2019objects,li2020generalized,ren2015faster,zhu2020deformable} rely on supervised learning with fixed data, where all classes are predefined and known beforehand. However, real-world data continuously evolve over time, leading to non-stationary distributions. Due to the \textit{stability-elasticity dilemma}, fine-tuning models directly on new class data leads to \textit{catastrophic forgetting} \cite{mccloskey1989catastrophic,mcrae1993catastrophic}, whereas joint training is expensive in both computation and storage. Therefore, incremental object detection (IOD) has attracted increasing attention in both research and practical applications.

Recent advances for IOD adopt knowledge distillation and exemplar replay to address forgetting. Knowledge distillation-based methods \cite{shmelkov2017incremental,li2019rilod,peng2021sid,feng2022overcoming,kang2023alleviating,peng2020faster} typically involve distillation on the non-background predictions of the old model to circumvent the imbalance issue caused by an excessive number of background predictions. Exemplar replay-based methods \cite{rebuffi2017icarl,joseph2021towards,liu2020multi,liu2023continual}, on the other hand, operate by preserving a small subset of samples from past training data (exemplars), replaying them in subsequent phases to mitigate the forgetting of old data. Despite the progress, there still exist limitations. First, due to the fixed model capacity, severe conflicts between preserving knowledge of old classes and incrementally learning new ones exist. This not only impedes the accommodation of new class knowledge but also leads to the overwriting and forgetting of old class knowledge. Second, exemplars with incomplete labels are inadequately equipped for replay. Even if pseudo labeling is an intuitive way, the low-quality supervision inherited from old knowledge will hamper the adaption.

To address the above concerns, we propose \textbf{Dy}namic object \textbf{Q}uery Assembly based \textbf{DE}tection \textbf{TR}ansformer, abbreviated as \textbf{DyQ-DETR}, for incremental object detection. Inspired by dynamic networks for incremental classification \cite{hung2019compacting,li2019learn,wang2022foster,yan2021dynamically,yoon2017lifelong,zhou2022model,douillard2022dytox} , we motivate to investigate the dynamic architecture for expanding model capacity in IOD. Specifically, the \textit{object queries} serve as the class-specific representative essence for memorization. The object queries from old classes can be assembled with learnable new queries for dynamically expandable representations with limited memory overhead and acceptable time overhead. As demonstrated in Fig.\ref{fig:intro}, we first obtain the visual features of the whole image with a CNN attached Transformer encoder. Then the set of object queries from previous steps are assembled with learnable ones corresponding to new classes in each incremental step. Moreover, we disentangle the self-attention and isolate the bipartite matching to remove the interaction between object queries from different steps. Besides, we propose the risk-balanced partial calibration to tackle incomplete annotations for effective exemplar replay.

Generally, we circumvent the \textit{catastrophic forgetting} in IOD from the perspective of dynamic networks, with the inspiration of incrementally adding class tokens has been proven effective in incremental learning \cite{zhou2022model,douillard2022dytox,shang2023incrementer, wu2023continual}. Notably, for object detection, multiple objects of a single class may co-exist in one image (sometimes even many). We adopt a many-to-many rather than one-to-one matching of dynamic queries, for which sparse object queries implicitly associate with the content and reference positional information of one or multiple seen old classes. We propose to memorize the set of object queries for sequentially arrived classes of data. An isolated group of queries is responsible for detecting the objects from a set of classes arriving at the same time step. By incrementally aggregating the category-wise object queries from previous steps and learning new class embedding, we decouple the representations for old and new class knowledge with lightweight query embeddings, simultaneously maintaining the stability and plasticity. As for exemplar replay, we propose to reserve images with moderate matching losses as exemplars, further perform partial calibration on the outputs of corresponding queries with only the annotated classes. Such risk-balanced partial calibration avoids over focusing on classes from any particular stage and eliminates the reliance on low-quality pseudo labels. Through dynamic object queries and risk-balanced partial calibration, our proposed DyQ-DETR significantly alleviates the forgetting in IOD.  % incremental object detection.
The main contributions are three-fold: 
% 1) We propose the dynamic object queries into Detection Transformer for incremental learning. In particular, the proposed DyQ-DETR is a simple yet effective way to expand the model capability compatible with new knowledge adaption and old knowledge preserving. \\
%2) We propose the disentangled self-attention for dynamic object queries. For dynamic knowledge expansion, the isolated bipartite matching over the object queries from different phases further decouples the representation learning of old classes and new ones. \\
% 3) A risk-balanced selection mechanism is presented for informative and reliable exemplars. The partial calibration is further utilized to tackle the incomplete annotations for effective exemplar replay.
\begin{itemize}[topsep=0pt, parsep=0pt, itemsep=0pt, partopsep=0pt, leftmargin=10pt]
\item We propose a novel approach to integrate dynamic object queries into Detection Transformer for incremental learning. By dynamically incorporating object queries into DETR, our approach provides a simple yet effective way to expand the model capability compatible with new knowledge adaption and old knowledge preserving. 
\item We propose the disentangled self-attention for dynamic object queries. For dynamic knowledge expansion, the isolated bipartite matching over the object queries from different phases further decouples the representation learning of old classes and new ones.
\item A risk-balanced selection mechanism is proposed to explore informative and reliable exemplars. The partial calibration is further associated to tackle the incomplete annotations for incremental detection with exemplar replay.
\end{itemize}

Extensive experiments on public benchmarks demonstrate the superiority of our proposed DyQ-DETR, outperforming the state-of-the-art methods by a large margin. It achieves the average \textbf{4.3\% $AP$} improvement in non-exemplar scenarios and \textbf{2.9\% $AP$} improvement with exemplar replay.

\section{Related Work}

\textbf{Incremental Learning (IL). }
Prevailing IL methods can be broadly divided as regularization-based, distillation-based and structure-based ones. Regularization-based methods\cite{aljundi2018memory,kirkpatrick2017overcoming,paik2020overcoming,zenke2017continual} estimate parameter importance and penalize updating of crucial parameters to maintain previous knowledge. Distillation-based methods build the mapping between the old and new model by matching logits\cite{li2017learning, rebuffi2017icarl}, feature maps\cite{douillard2020podnet}, or other information\cite{joseph2022energy, pourkeshavarzi2021looking, simon2021learning, tao2020topology, wang2022foster}, which leverage the knowledge transfer to prevent forgetting. Structure-based methods\cite{hung2019compacting,li2019learn,wang2022foster,yan2021dynamically,yoon2017lifelong,zhou2022model} dynamically expand the representative network, \textit{e.g.,} backbone, prompt, to fit the evolving data stream.

% \subsection{Incremental Object Detection}
\textbf{Incremental Object Detection (IOD). } As the typical extension of incremental learning, IOD involves multiple objects belonging to the old and new classes appearing simultaneously. This co-occurrence makes knowledge distillation an inherently effective strategy for IOD, since it allows for the utilization of old class objects from new training samples to minimize the discrepancies in responses between the previous and current updating model. As a pioneering work, ILOD\cite{shmelkov2017incremental} distills the responses for old classes to counteract catastrophic forgetting on Fast R-CNN\cite{girshick2015fast}. The idea of knowledge distillation is then extended to other detection frameworks, such as CenterNet\cite{zhou2019objects} (SID\cite{peng2021sid}), RetinaNet\cite{lin2017focal} (RILOD\cite{li2019rilod}), GFLV1\cite{li2020generalized} (ERD\cite{feng2022overcoming}), Faster R-CNN\cite{ren2015faster} (CIFRCN\cite{hao2019end}, Faster ILOD\cite{peng2020faster}, DMC\cite{zhang2020class}, BNC\cite{dong2021bridging}, IOD-ML\cite{joseph2021incremental}) and Deformable DETR\cite{zhu2020deformable} (CL-DETR\cite{liu2023continual}). Built on Deformable DETR instead of conventional detectors such as Faster R-CNN, DyQ-DETR can efficiently expand queries rather than inefficiently enlarging the backbone or specific convolutional layers. Note that DyQ-DETR also uses knowledge distillation techniques. 
As for exemplar replay,\cite{joseph2021towards} proposes maintaining an exemplar set and fine-tuning the model on the exemplars after each incremental step.\cite{liu2020multi} proposes an adaptive sampling strategy to achieve more efficient exemplar selection and \cite{liu2023continual} proposes distribution-preserving calibration, which selects exemplars to match the training distribution. They usually finetune directly using the exemplar set with incomplete annotations and overlook the amount of information and reliability of the annotated objects. % These two issues can be resolved by our proposed partial calibration and risk-balanced exemplar selection mechanism, respectively.

% \subsection{Transformer-based Object Detection}

\textbf{Transformer-based Object Detection.}
The infusive work DETR (DEtection TRansfomer)\cite{carion2020end} formulates object detection as a set prediction problem, with an elegant transformer-based architecture. It captures global context and reasons object relations with attention mechanism. With a small set of learnable object queries and Hungarian bipartite matching\cite{kuhn1955hungarian}, it eliminates the need for the complex non-maximum suppression and many other hand-designed components in object detection while demonstrating good performance. Deformable DETR\cite{zhu2020deformable} introduces sparse attention on multi-level feature maps, thereby accelerating the convergence of DETR and improving the performance, particularly for small objects. There also exists many other DETR variants\cite{dai2021up,meng2021conditional,li2022dn,liu2022dab} designed to accelerate convergence speed and enhance detection performance. Without loss of generality, we build our method on the commonly used Deformable DETR.

\section{Methodology}

\noindent\textbf{Preliminaries}
In the paradigm of IOD, object detection is performed in multiple steps from sequentially arrived training data to recognize and localize objects of all seen classes in test images. Let $D$ be a dataset with samples $(x,y)$, where $x$ is the image, $y$ is the set of object class labels and the associated bounding boxes. Suppose there are $T$ steps. At time step $t$, the incoming dataset is denoted as $D_t=(x_t, y_t)$ and the objects belong to seen classes in $C_t$. Specially, the images in $D_t$ are only annotated for objects in $C_t$. Moreover, the class sets at different time steps are mutually exclusive from each other, \textit{i.e.}, $C_t\cap C_{\tau}=\varnothing$, $t\neq \tau$. Since $D_t$ only contains the annotations with classes in $C_t$, each class can be learned only once in a specific step. Due to the missing annotations of old classes $\{C_1, C_2, \cdots, C_{t-1}\}$ in current data $D_t$, detection models are prone to forget the knowledge of old classes and bias towards new ones. To overcome  the catastrophic forgetting, we introduce the dynamic object queries into Detection Transformer and propose DyQ-DETR for IOD.

We construct the DyQ-DETR based on Detection Transformer for incremental object detection. In particular, we elaborate the details of dynamic object queries with the structure design and training strategy in Sec. \ref{method:query}. The risk-balanced exemplar selection and partial calibration are in Sec. \ref{method:risk}.

\begin{figure}[tb]
  \centering
  \includegraphics[height=5.2cm]{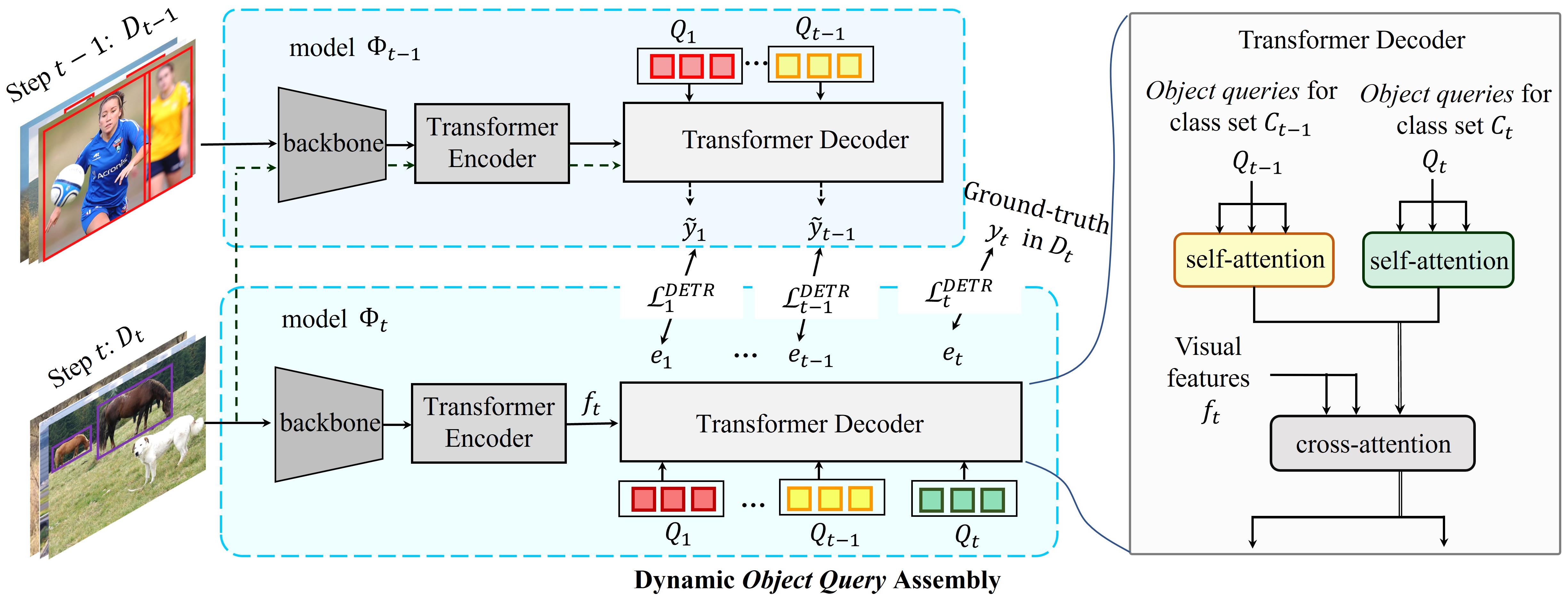}
  \caption{The overview of our proposed DyQ-DETR. The dynamic object queries serve as the input for Transformer decoder in incremental learning. At each incremental time step $t$, for an image $x \in D_t$, the training loss is independently computed. The total loss is the weighted sum of the knowledge distillation loss $\mathcal{L}_i^{DETR}(1 \leq i < t)$ caused by pseudo labels and the standard DETR loss $\mathcal{L}_t^{DETR}$ caused by ground-truth labels. 
  }
  \label{fig:framework}
\end{figure}

\subsection{Transformer-based Incremental Object Detection}

We build up the DyQ-DETR based on the DETR architecture~\cite{carion2020end,zhu2020deformable}. Other than the backbone, a DETR consists of an encoder, a Transformer decoder, and the predictor to generate object classes and locations. The encoder takes the images as inputs and outputs visual features. Then those visual features and learnable object queries are fed into Transformer decoder for prediction. Notably, we propose to aggregate the object queries from up-to-far learning steps to resist the forgetting of old class knowledge. Due to the absence of annotations from previous seen classes in incremental step $t$, knowledge distillation is applied for preserving the class-specific old knowledge. As in Fig.\ref{fig:framework}, we select the non-background predictions of old classes by thresholding. Pseudo labels $\tilde{y}_{\tau} (1 \leq \tau<t)$ are generated by the last previous model. Instead of mixing the pseudo labels $\tilde{y}_{\tau} (1 \leq \tau<t)$ with the ground-truth labels $y_t$, we use $\tilde{y}_{\tau}$ and $y_t$ separately to supervise the model training. The pseudo labels for old classes and the real ground-truths are used to guide the learning of object queries being $Q_\tau$ and $Q_t$, respectively. Besides, object queries from different groups share the weight parameters of the Transformer decoder. To allow for computational complexity to grow linearly rather than quadratically, the self-attention over different groups of object queries are eliminated.

As for IOD with exemplar replay, where a small number of exemplar images $\epsilon_t$ from dataset $D_t$ in different time steps are stored, we introduce a risk-balanced selection mechanism. At time step $t$, we use the trained model $\Phi_t$ to score the annotated objects from images in $D_t$. The computed loss from the partial bipartite matching is considered as the risk score for exemplar selection. We choose the sample images with moderate risk score for the tradeoff between the annotation importance and quality. Specifically, to build the exemplar dataset $\epsilon_{1:t}$, we select the samples falling into the middle part after sorting, because they are informative and reliably annotated. Considering the image in $\epsilon_{1:t}$ is incompletely annotated for specific classes, we adapt the partial calibration. We leverage the incomplete annotations to calibrate the output for corresponding object queries in each group. Since $\epsilon_{1:t}$ is balanced, the partial calibration will prohibit the model being biased towards certain classes.

\subsection{Dynamic Object Query Assembly}
\label{method:query}

Existing DETR models employ a fixed set of object queries (\textit{i.e.,} learnable embedding) as the inputs of Transformer decoder. These object queries are progressively optimized to map into object instances in images. Despite various designs, the learnable object queries are highly relevant with the specific classes. For IOD, the object queries are expected to be associated with objects belonging to sequentially arrived classes. Since new classes are considered as backgrounds in previous time steps, the preservation of old knowledge naturally contradicts the knowledge updating from new data learning, especially from the perspective of object queries. Moreover, the conflict between a fixed network and the continually emerged class-specific information severely undermines the performance of incremental learning, especially in non-exemplar scenarios.

To cope with incremental classes without extra modules in network architecture, we focus on tackling the forgetting issue with dynamic object queries. At time step $t$, for a set of new classes $C_t$, a new set of learnable object queries $Q_t$ is added. The newly added object queries $Q_t$ are aggregated with previous sets of queries $Q_{\tau}, 1 \leq \tau<t$. The assembly of object queries $\{Q_1, Q_2, \cdots, Q_t\}$ serves as the input of Transformer encoder in step $t$. The visual embeddings $e_t$ corresponding to the newly expanded queries $Q_t$ are used to predict the objects of new classes $C_{t}$, and the old classes $C_{\tau} (1 \leq \tau<t)$ is detected with the embeddings $e_{\tau}$ of old queries $Q_{\tau}$. By dynamically expanding object queries, the new and old classes are segregated by class embeddings, significantly alleviating the conflict between the old knowledge and continuously evolving new knowledge.

Based on the dynamic assembly of object queries, we further investigate the decoder design to restrict the computational burden. In standard DETR, object queries interact visual features with cross-attention for refinement. Besides, those object queries interact with each other by self-attention. 
Through self-attention, duplicated detections can be removed, but the computational complexity increases quadratically with the number of object queries. Considering that the object instances from different class sets rarely overlap, we disentangle the self-attention in Transformer decoder. The self-attention is computed among separate set of object queries, as:
\begin{equation}
    a_{i,j}=
    \begin{cases}
    Softmax(Q_{m,i}\cdot Q_{n,j}^T/\sqrt{d})& {m=n}\\
    0& {m\neq n}
    \end{cases},
    \label{eq:attention}
\end{equation}
where $a_{i,j}$ denotes the attention weight between two queries $Q_{m,i}$ and $Q_{n,j}$ from the query set $Q_{m}$ and $Q_{n}$, respectively. By eliminating the attention interaction between queries of different groups as shown in Eq.(\ref{eq:attention}), we achieve a linear growth in computational complexity almost for free. Upon the addition of new queries, the capability of old queries to detect old classes is fully preserved. We perform the decoder forward passes as many as the time steps, obtaining the embedding vectors from different sets of queries as: 
\begin{equation}
  e_{\tau}=decoder(f_t, Q_{\tau}), 1 \leq \tau \leq t,
  \label{eq:decoder}
\end{equation}
where $f_t$ denotes the visual feature extracted from image $x$ by the CNN and Transformer encoder. Each decoder forward pass is executed with a different set of queries $Q_{\tau}$, resulting in different task-specific embeddings $e_{\tau}$ to obtain detection predictions of the corresponding classes $C_{\tau}$.

We further adapt the knowledge distillation for incremental detector training. The foreground predictions with the pseudo labels generated from old model are kept for distillation and used for supervision. As in \cite{liu2023continual}, the foreground predictions with high confidence from the old model are selected. A probability threshold $\theta_p$ (typically 0.4) is set over the prediction scores. An additional IoU threshold $\theta_{IoU}$ (typically 0.7) is used to restrict the predictions not too close to the ground-truth bounding boxes of new classes. It helps filter out incorrect predictions about new class objects that are misclassified as old classes. The highly-confident predictions after filtering are used as pseudo labels $\tilde{y}_{\tau} (1 \leq \tau < t)$, which contain two parts of annotations (\textit{i.e.,} the predicted object labels and bounding boxes). Notably, 
instead of merging the pseudo labels with real grounding truths, we compute the separate bipartite matching losses with different set of object queries in an independent way. The loss for retaining old class knowledge $\mathcal{L}_{\tau}^{DETR} (1 \leq \tau < t)$, as well as the loss for learning new class knowledge $\mathcal{L}_t^{DETR}$ can be formulated as in Eq.(\ref{eq:loss1}):
\begin{equation}
  \mathcal{L}_{\tau}^{DETR} = \mathcal{L}^{DETR}(e_{\tau},\tilde{y}_{\tau}), 1 \leq {\tau} < t ; 
  \mathcal{L}_t^{DETR} = \mathcal{L}^{DETR}(e_{t}, y_t), \label{eq:loss1}
\end{equation}
Note that the specific embeddings $e_{\tau}, 1 \leq \tau \leq t$ are only supervised with the corresponding pseudo labels or real annotations, leading to decoupled representations. The total loss is the weighted sum as:
\begin{equation}
  \mathcal{L}_{total}=\sum_{\tau=1}^{t}w_\tau\mathcal{L}_\tau^{DETR}.
  \label{eq:total_loss}
\end{equation}
To tackle the varying number of classes in each step and prevent the model from being biased towards class sets with fewer classes, we set the weight $w_{\tau}$ of the class set $C_{\tau}$ to ${|C_{\tau}|}/{|C_{1:t}|}$.

\begin{wrapfigure}{r}{0.5\textwidth}
  \centering
  \vspace{-10mm}
  \includegraphics[height=5.2cm]{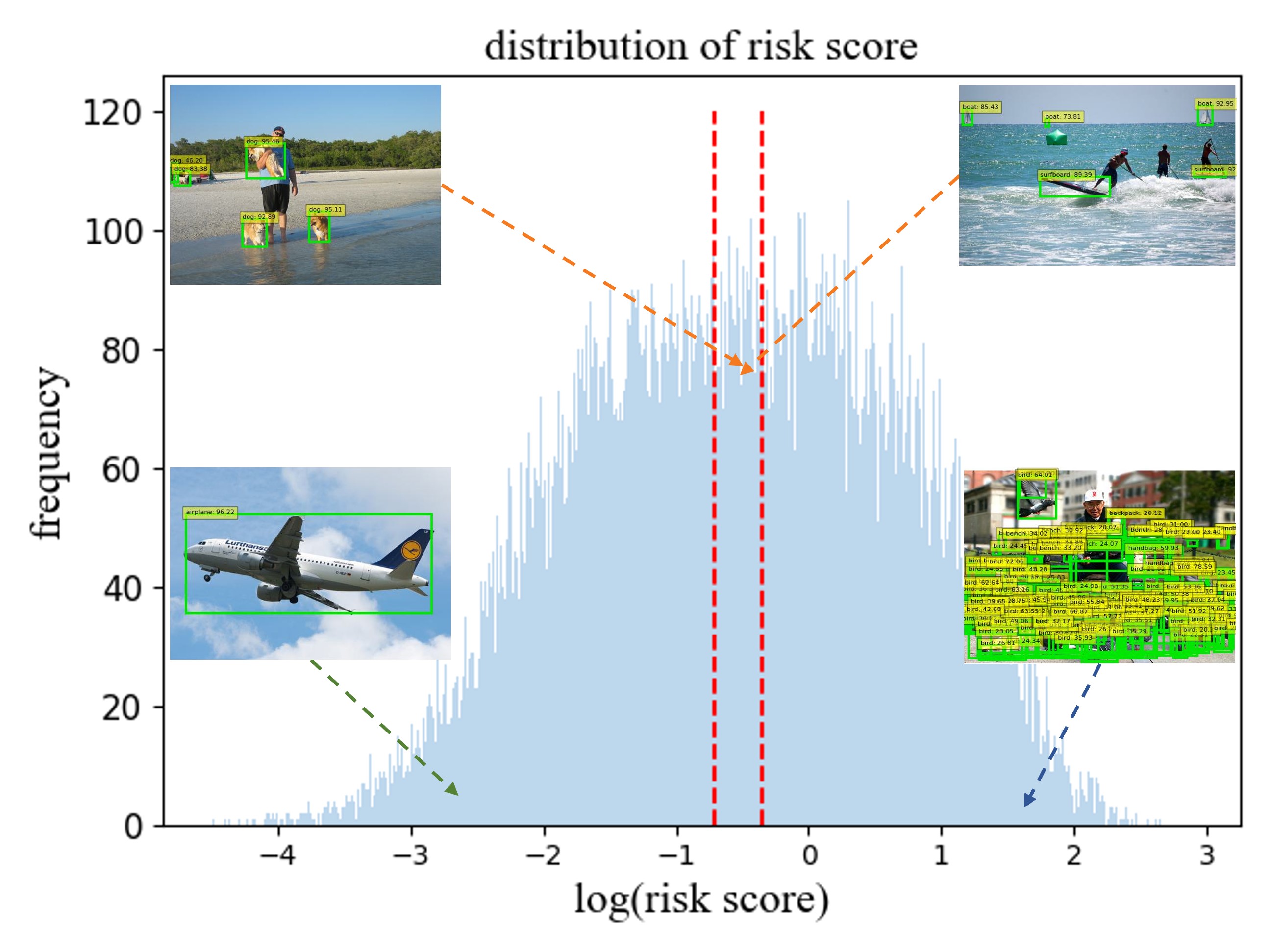}
  \caption{Illustration of risk-balanced exemplar selection. We choose the middle part with moderate risk score to serve as the exemplars.}
  \vspace{-3mm}
  \label{fig:risk}
\end{wrapfigure}

\subsection{Risk-balanced Partial Calibration}
\label{method:risk}

For exemplar replay based IOD paradigm, an exemplar memory is formed to store a small number of samples for subsequent incremental learning. At time step $t$, the exemplar set $\epsilon_t$ is a subset of the entire dataset $D_t$. Let $\epsilon_{1:t}=\epsilon_1 \cup \cdots \epsilon_{t}$. The exemplars in $\epsilon_t$ are used to represent the samples with objects in $C_t$. We intend to choose the class annotated images which are \textit{substantial} for detector training. An intuitive way is to directly compute the loss between the model output and the real label $y_t$ for sample selection. However, since the image $x \in D_t$ only contains annotations for the classes $C_t$ of interest, this will result in the loss being dominated by the absence of old classes, posing challenges to foreground-background balancing. 

Benefiting from the internal decoupling of dynamic object queries, we are able to exclusively detect new classes $C_t$ using the corresponding queries $Q_t$. Considering that the image is only annotated for the specific classes, the partial loss is more reliable. For computing the partial loss, the old classes are considered as backgrounds by $Q_t$, which is compatible with the incomplete real labels $y_t$. After each incremental step of training, the partial loss from Eq.(\ref{eq:partial_loss}) is considered as the risk score to guide the subsequent selection of exemplars.
\begin{equation}
  \Upsilon \leftarrow \mathcal{L}_{partial}=\mathcal{L}_t^{DETR}(e_t, y_t).
  \label{eq:partial_loss}
\end{equation}

% \begin{figure}[tb]
%   \centering
%   \includegraphics[height=5.6cm]{images/risk_nips_1.jpg}
%   \caption{Illustration of risk-balanced exemplar selection. We simply use the partial training loss with real ground truths as the risk. Images with low risk provide limited information for model optimization, while those with high risk are likely to contain crowded false positives with low-quality annotations. We choose the middle part with moderate risk to serve as the exemplars.}
%   \label{fig:risk}
% \end{figure}

The risk score can measure the quality of incomplete class labels and bounding boxes. Additionally, it also takes the number of annotated objects like in \cite{liu2023continual} into account. As in Fig.~\ref{fig:risk}, images with low risk, which constitute a high proportion, provide limited information for optimization, while those with high risk are likely to be outliers with incorrect annotations. Based on the risk estimation, we construct the exemplar set $\epsilon_t$ by sorting and selecting the middle part of risk-balanced samples in $D_t$. The exemplar set $\epsilon_t$ is merged with $\epsilon_{t-1}$ to form a set $\epsilon_{1:t}$ for partial calibration of the model.

In incremental step $t$, the exemplar sets $\epsilon_{1:t}$ are used to finetune the model after training with $D_t$. Previous IOD methods typically finetune the model directly with dataset $\epsilon_{1:t}$, without addressing the issue of missing labels. That is, an image $x$ in the balanced exemplar set $\epsilon_{1:t}$ is only annotated for a specific class subset from $\{C_1, ...,C_{t-1}, C_t\}$, with the absence of annotations for other classes. The confusing, even contradictive supervision hinders the process of prediction calibration. An intuitive way is to use pseudo labels, yet the quality of pseudo labels is hard to guarantee. Thanks to the dynamic object queries and associative disentangled computation, we propose to perform partial calibration that relies only on the incomplete real labels. Specifically, we calculate the partial loss between the outputs of the corresponding queries and the ground-truth annotations in $\epsilon_{1:t}$. This type of partial calibration further mitigates the forgetting.

\section{Experiment}
\label{sec:Experiments}

% In this section, we compare our DyQ-DETR with state-of-the-art (SOTA) methods on different IOD settings and we perform detailed ablations to validate the effectiveness of DyQ-DETR components. 

\subsection{Experimental Setup}

\noindent\textbf{Dataset and evaluation metrics.} Following\cite{liu2023continual}, we conduct experiments on the widely-used COCO 2017 dataset\cite{lin2014microsoft}, which consists of images from 80 object categories in natural scenes. The standard COCO metrics as AP, AP$_{50}$, AP$_{75}$, AP$_{S}$, AP$_{M}$, AP$_{L}$ are used for performance evaluation.
% All experiments are based on COCO 2017 dataset, and we will not specify this in subsequent descriptions. 

\noindent\textbf{Protocols.} We evaluate DyQ-DETR with two protocols: \textit{1)} traditional protocol \cite{shmelkov2017incremental} (Tab.\ref{tab:main_comp}-left) and \textit{2)} revised protocol proposed by\cite{liu2023continual} (Tab.\ref{tab:main_comp}-right). Protocol \textit{2)} avoids observing the same images at different stages. Therefore, we adopt protocol \textit{2)} for all subsequent experiments, and the details of protocol \textit{1)} can be found in the appendix. For protocol \textit{2)}, we adopt both two-phase and multiple-phase settings, which can be formulated in the form of $c_1+c_2+...+c_T$, where $c_t$ represents the number of new classes in incremental step $t$ ($c_t=|C_t|$) and the sum of $c_t$ is denoted as $c$ ($c=|C_{1:T}|$). At time step $t$, we observe a fraction $\frac{c_t}{c}$ of the training samples with $c_t$ new categories annotated. We test settings $c_1+c_2+...+c_T$ = $40+40$, $70+10$, $40+20 \times 2$, and $40+10 \times 4$. Following \cite{liu2023continual}, we also set the total memory budget for the exemplars to be 10\% of the total dataset size.

\begin{table}[t]
    \caption{IOD results (\%) on COCO 2017 under the 40+40 and 70+10 scenarios. $\dagger$ indicates that the results are obtained without exemplar replay. “Upper bound” refers to the result of joint training with all previous data available at each step.} 
    \label{tab:main_comp}
    \begin{subtable}{.5\linewidth}
    \centering
  \small
  \scalebox{0.87}{
  \begin{tabular}{p{8pt}|p{73pt}p{8pt}p{9pt}p{9pt}p{8pt}p{8pt}p{8pt}}% {l|lllllllll}
    \toprule
      \textit{1)}  & Method     & \footnotesize{AP}     & AP$_{50}$     & AP$_{75}$     & AP$_{S}$     & AP$_{M}$     & AP$_{L}$ \\
    \midrule
    \multirow{8}{*}{\rotatebox{90}{40+40}}
    &\footnotesize{LwF\cite{li2017learning}$\dagger$}  & 17.2  & 25.4  & 18.6  & 7.9  & 18.4  & 24.3  \\
    &\footnotesize{CL-DETR\cite{liu2023continual}$\dagger$}  &39.2   &56.1   &42.6   &21.0   &42.8   &52.6   \\
    &\footnotesize{DyQ-DETR (ours) $\dagger$}   & \textbf{41.4}  & \textbf{59.7}  & \textbf{44.9}  & \textbf{24.1}  & \textbf{45.2}  & \textbf{54.3}  \\
    \cmidrule(r){2-8}
    &\footnotesize{RILOD\cite{li2019rilod}}  & 29.9  & 45.0  & 32.0  & 15.8  & 33.0  & 40.5  \\
    &\footnotesize{SID\cite{peng2021sid}}  & 34.0  & 51.4  & 36.3  & 18.4  & 38.4  & 44.9  \\
    &\footnotesize{ERD\cite{feng2022overcoming}}  & 36.9  & 54.5  & 39.6  & 21.3  & 40.4  & 47.5  \\
    &\footnotesize{CL-DETR\cite{liu2023continual}}  &42.0   &60.1   &45.9   &24.0   &45.3   &55.6   \\
    &\footnotesize{DyQ-DETR (ours)}   & \textbf{42.4}  & \textbf{60.5}  & \textbf{45.9}  & \textbf{23.9}  & \textbf{46.3}  & \textbf{56.7}  \\
   
    \midrule
    \multirow{8}{*}{\rotatebox{90}{70+10}}
    & \footnotesize{LwF\cite{li2017learning}$\dagger$}  & 7.1  & 12.4  & 7.0  & 4.8  & 9.5  & 10.0  \\
    & \footnotesize{CL-DETR\cite{liu2023continual}$\dagger$}  &35.8   &53.5   &39.5   &19.4   &41.5   &46.1   \\
    & \footnotesize{DyQ-DETR (ours)$\dagger$ }   & \textbf{39.5}  & \textbf{56.4}  & \textbf{43.1}  & \textbf{22.5}  & \textbf{43.1}  & \textbf{53.0}  \\
    \cmidrule(r){2-8}
    & \footnotesize{RILOD\cite{li2019rilod}}   & 24.5  & 37.9  & 25.7  & 14.2  & 27.4  & 33.5   \\
    & \footnotesize{SID\cite{peng2021sid}}  & 32.8  & 49.0  & 35.0  & 17.1  & 36.9  & 44.5   \\
    & \footnotesize{ERD\cite{feng2022overcoming}}  & 34.9  & 51.9  & 37.4  & 18.7  & 38.8  & 45.5   \\
    & \footnotesize{CL-DETR\cite{liu2023continual}}   &40.4   &58.0   &43.9   &23.8   &43.6   &53.5     \\
    & \footnotesize{DyQ-DETR (ours)}   & \textbf{42.4}  & \textbf{60.4}  & \textbf{46.3}  & \textbf{24.5}  & \textbf{45.7}  & \textbf{57.5}  \\
    
    \bottomrule
  \end{tabular}}

    \end{subtable}%
    \begin{subtable}{.5\linewidth}
     
     \centering
  \small
  \scalebox{0.83}{
  \begin{tabular}{p{8pt}|p{73pt}p{8pt}p{9pt}p{9pt}p{8pt}p{8pt}p{8pt}}% {l|lllllllll}
    \toprule
        \textit{2)} & Method     & \footnotesize{AP}     & AP$_{50}$     & AP$_{75}$     & AP$_{S}$     & AP$_{M}$     & AP$_{L}$ \\
    \midrule
    \multirow{8}{*}{\rotatebox{90}{40+40}}
    & \footnotesize{Upper Bound}  &41.0 &59.8 &44.2 &25.0 &43.5 &54.2  \\
    \cmidrule(r){2-8}
    & \footnotesize{LwF\cite{li2017learning}$\dagger$}  & 23.9  & 41.5  & 25.0  & 12.0  & 26.4  & 33.0  \\
    & \footnotesize{CL-DETR\cite{liu2023continual}$\dagger$}  & 36.2  & 52.6  & 39.5  & 18.7  & 39.5  & 49.4  \\
    & \footnotesize{DyQ-DETR (ours) $\dagger$}  & \textbf{39.1} & \textbf{57.1}  & \textbf{42.5}  & \textbf{21.3}  & \textbf{42.7}  & \textbf{51.8}  \\
    \cmidrule(r){2-8}
    & \footnotesize{iCaRL\cite{rebuffi2017icarl}}  & 33.4  & 52.0  & 36.0  & 18.0  & 36.4  & 45.5  \\
    & \footnotesize{ERD\cite{feng2022overcoming}}  & 36.0  & 55.2  & 38.7  & 19.5  & 38.7  & 49.0  \\
    & \footnotesize{CL-DETR\cite{liu2023continual}}  & 37.5  & 55.1  & 40.3  & 20.9  & 40.8  & 50.7  \\
    & \footnotesize{DyQ-DETR (ours)}  & \textbf{39.7}  & \textbf{57.5}  & \textbf{43.0}  & \textbf{21.6}  & \textbf{42.9}  & \textbf{53.8}  \\
    \midrule
    \multirow{8}{*}{\rotatebox{90}{70+10}}
    & \footnotesize{Upper Bound}  &43.3 &61.8 &47.0 &25.3 &46.1 &57.9  \\
    \cmidrule(r){2-8}
    & \footnotesize{LwF\cite{li2017learning}$\dagger$}  & 24.5  & 36.6  & 26.7  & 12.4  & 28.2  & 35.2  \\
    & \footnotesize{CL-DETR\cite{liu2023continual} $\dagger$}  & 34.0  & 48.0  & 37.2  & 15.5  & 37.7  & 49.7  \\
    & \footnotesize{DyQ-DETR (ours) $\dagger$}  & \textbf{39.6}  & \textbf{57.6}  & \textbf{43.5}  & \textbf{23.4}  & \textbf{43.3}  & \textbf{51.8}  \\
    \cmidrule(r){2-8}
    & \footnotesize{iCaRL\cite{rebuffi2017icarl}}  & 35.9  & 52.5  & 39.2  & 19.1  & 39.4  & 48.6   \\
    & \footnotesize{ERD\cite{feng2022overcoming}}  & 36.9  & 55.7  & 40.1  & 21.4  & 39.6  & 48.7   \\
    & \footnotesize{CL-DETR\cite{liu2023continual}}  & 40.1  & 57.8  & 43.7  & 23.2  & 43.2  & 52.1    \\
    & \footnotesize{DyQ-DETR (ours)} & \textbf{41.9}  & \textbf{60.1}  & \textbf{45.8}  & \textbf{24.1} & \textbf{45.3}  & \textbf{55.8} \\
    \bottomrule
  \end{tabular}}

    \end{subtable}
    \label{main_table}
\end{table}

\noindent\textbf{Implementation details.} Following \cite{liu2023continual}, we build DyQ-DETR on top of Deformable DETR \cite{zhu2020deformable} without iterative bounding box refinement and the two-stage variant. The backbone is ResNet-50\cite{he2016deep} pretrained on ImageNet\cite{deng2009imagenet} and the training configurations for the initial stage are consistent with\cite{liu2023continual} to maintain uniform performance in the initial phase. We denote the initial number of queries for the detector as $N$ (typically 300). For both two-phase and multi-phase settings, we dynamically expand by $N$ queries at each incremental step $t$ and the initial parameters of $Q_t$ are inherited from $Q_{t-1}$. At step $t$, old queries $Q_{1:t-1}$ are frozen during the incremental training and unfrozen in subsequent exemplar replay. 
We train the model for 50 epochs, and for an additional 20 epochs during fine-tuning. All the experiments are performed on 8 NVIDIA GeForce RTX 3090, with a batch size of 8.

\subsection{Quantitative Results}

\noindent\textbf{Two-phase setting.} We compare our DyQ-DETR with LwF\cite{li2017learning}, iCaRL\cite{rebuffi2017icarl}, RILOD\cite{li2019rilod}, SID\cite{peng2021sid}, ERD\cite{feng2022overcoming}, and the previous SOTA method CL-DETR\cite{liu2023continual}. For each setting, we provide the performance of different methods with/without Exemplar Replay (ER). The metrics by joint training are also presented as the upper bound for reference. Tab.\ref{tab:main_comp} shows that, in the two-phase settings, our proposed DyQ-DETR consistently outperforms the aforementioned methods under different protocols with significant margins. For protocol \textit{2)}, with exemplar replay, the DyQ-DETR achieves the $AP$ of 39.7\% and 41.9\% under 40+40 and 70+10 settings. It surpasses CL-DETR by 2.2\% $AP$ and 1.8\% $AP$ under the 40+40 and 70+10 settings, respectively. Compared with the upper bound, the DyQ-DETR obtains an average performance gap of 1.4\%, which is much smaller than the 3.4\% gap of CL-DETR. To evaluate the model's ability to preserve old knowledge and learn new knowledge, we conduct a comparison with CL-DETR in the 40+40 setting, where both capabilities are equally important. CL-DETR achieves an old class $AP$ of 39.7\% and a new class $AP$ of 36.3\%, while our method achieves much better results with 41.3\% $AP$ for old classes and 38.6\% $AP$ for new classes, respectively. This validates that our proposed method achieves better stability and plasticity, thus addressing the catastrophic forgetting effectively.

\begin{figure}[tbp]
  \centering
  \includegraphics[width=13.5cm, height=3.8cm]{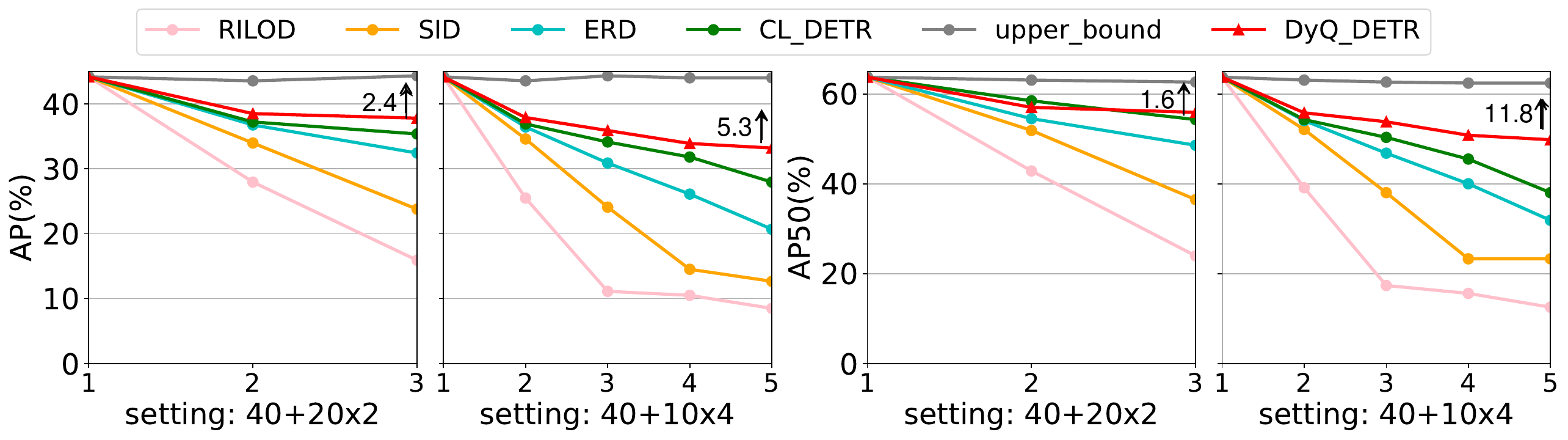}
  \caption{IOD results ($AP/AP_{50}, \%$) in the multi-phase 40+20×2 and 40+10×4 settings. The results of all other works are from \cite{liu2023continual}. % (b) A comparison of parameter (top) and complexity overhead (bottom) at each stage. We conduct experiments in the challenging 40+10×4 setting, with each expansion adding $N$ queries ($N$=300). DSA denotes disentangled self-attention.
  }
  \label{fig:multi_phase}
\end{figure}

Additionally, Tab.\ref{tab:main_comp} include a performance comparison without exemplar replay (\textit{w/o} ER) to show the effectiveness of dynamic object queries. In non-exemplar scenarios, DyQ-DETR demonstrates a more significant advantage, outperforming CL-DETR by 2.9\% $AP$ and 5.6\% $AP$ in the 40+40 and 70+10 settings, respectively. It is noteworthy that the performance of our method without ER is comparable or even exceeds that of the existing method with ER. For example, the performance of our DyQ-DETR \textit{w/o} ER is 1.6\% $AP$ higher than that of CL-DETR \textit{w/} ER in the 40+40 setting.

\noindent \textbf{Multiple-phase setting.} We conduct experiments in the more challenging 40+20$\times$2 and 40+10$\times$4 settings. The changing $AP$ and $AP_{50}$ along with time steps are presented in Fig.\ref{fig:multi_phase}. Our DyQ-DETR consistently outperforms other IOD methods. Moreover, in both settings, the $AP$ improvements of DyQ-DETR become more pronounced with the increase of incremental steps.

\begin{wrapfigure}{r}{0.5\textwidth}
  \centering
  \vspace{-2mm}
  \includegraphics[width=7.0cm, height=2.8cm]{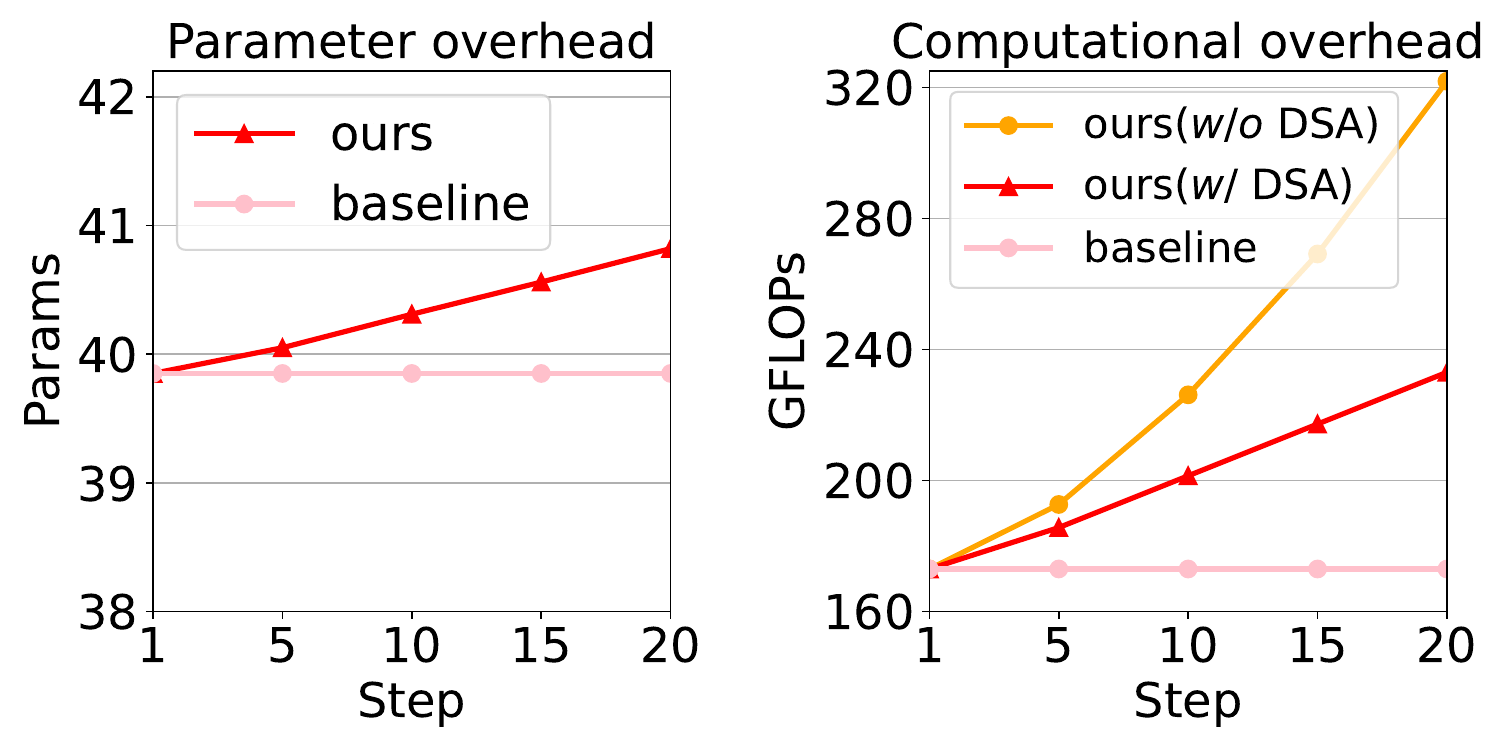}
  % \vspace{-3mm}
  \caption{Comparison of parameter (left) and complexity overhead (right) with the addition of 100 queries at each step. DSA denotes disentangled self-attention.}
  \vspace{-2mm}
  \label{fig:overhead}
\end{wrapfigure}

\noindent \textbf{Scalability.} As shown in Fig.\ref{fig:overhead}-left, by expanding queries rather than the network structure, the additional parameter overhead of our method is almost negligible. As illustrated in Fig.\ref{fig:overhead}-right, the computational overhead of our method grows linearly because of the removed inter-group query interaction. Since the computational load of the decoder (excluding the portion shared by different query groups) constitutes a small portion ($\sim$6\%) of the entire model's computation, our computational complexity increases at a slow linear rate. Specifically, with 20 stages and an increment of 100 queries per stage (note that the standard Deformable DETR has 300 queries), DyQ-DETR only increases the parameters and GFLOPs by 2\% and 39\% respectively compared to the standard Deformable DETR, confirming its scalability.

% Besides, since the computational load of the decoder constitutes a relatively small portion of the entire model's computation, our computational complexity increases at a slow linear rate.

\subsection{Ablation Study}

\noindent\textbf{Effect of dynamic object queries.} Tab.3 illustrates the ablation study over dynamic object queries in the 70+10 setting. Compared to the baseline, the naive way of expanding queries (+Nat Query) increases the $AP$ to 34.5\% and $AP_{50}$ to 52.7\%, by 1.1\% and 4.2\%, respectively. This can be attributed to model capacity. By equipping the dynamic object queries (+Dy Query) with isolated matching on independent query set, we further obtain 3.6\% $AP$ and 3.5\% $AP_{50}$ improvements. It validates that dynamic object queries is effective to retain old knowledge during the learning of new knowledge. Furthermore, freezing the task-specific old queries (+DyFro Query) during the incremental training leads to slightly more improvement of 0.4\% $AP$.

\noindent\textbf{Components for exemplar replay.} We introduce the risk-balanced exemplar selection (RS) and partial calibration (PC) for exemplar replay. The ablative results of the two components in the 70+10 setting are provided in Tab.\ref{tab:ablation2}. We compare the RS and PC with the distribution-preserving selection (DS) and direct calibration (DC) proposed in CL-DETR~\cite{liu2023continual}, respectively. 
By separately applying RS and PC, we increase the $AP$ by 0.5\% and 0.9\%, respectively. Applying the whole risk-balanced exemplar selection and partial calibration obtains 41.9\% $AP$, exceeding the baseline by 1.5\%.

\begin{table*}[t]
\begin{floatrow}
% \capbtabbox{
% \scalebox{0.77}{
% \begin{tabular}{ccc|ccccccc}
% \toprule
%   DyQ & Dep & Frozen & $AP$     & $AP_{50}$     & $AP_{75}$     & $AP_{S}$     & $AP_{M}$     & $AP_{L}$       \\
% \midrule
% % 1&& & & &4.2 &6.4 &4.5 &1.9 &4.5 &5.7  \\
%   & & & 34.5  & 48.5  & 37.7  & 16.7  & 37.4   & 49.8  \\
%  \checkmark & & &35.6 &52.7 &38.5 &20.3 &37.8 &46.4  \\
%  \checkmark & \checkmark & &39.2 &56.2 &43.1 &21.6 &42.8 &52.1  \\
%  \checkmark & \checkmark & \checkmark & 39.6  & 57.6  & 43.5  & 23.4  & 43.3  & 51.8  \\
% \bottomrule
% \end{tabular}}
% }{
%  \caption{.}
%  \label{tab:ablation1}
%  \small
% }
\hspace{-0.2in}
% \setcaptionwidth{2.7in}
\capbtabbox{
\scalebox{0.7}{
\begin{tabular}{l|ccccccc}
\toprule
  & $AP$     & $AP_{50}$     & $AP_{75}$     & $AP_{S}$     & $AP_{M}$     & $AP_{L}$       \\
\midrule
% 1&& & & &4.2 &6.4 &4.5 &1.9 &4.5 &5.7  \\
 Baseline & 34.5  & 48.5  & 37.7  & 16.7  & 37.4   & 49.8  \\
 +Nat Query &35.6 &52.7 &38.5 &20.3 &37.8 &46.4  \\
 +Dy Query  &39.2 &56.2 &43.1 &21.6 &42.8 &52.1  \\
 +DyFro Query & 39.6  & 57.6  & 43.5  & 23.4  & 43.3  & 51.8  \\
\bottomrule
\end{tabular}}
}{
 \caption{Ablation on dynamic object queries. }
 \label{tab:ablation1000}
 \small
}

\hspace{-0.3in}
% \setcaptionwidth{2.2in}
\capbtabbox{
\scalebox{0.65}{
\begin{tabular}{cc|cc|cccccc}
\toprule
  DS      & RS & DC      & PC       & $AP$     & $AP_{50}$     & $AP_{75}$     & $AP_{S}$     & $AP_{M}$     & $AP_{L}$       \\
\midrule
\checkmark& $\times$ & \checkmark &$\times$ &40.5 &58.7 &44.5 &24.7 &44.0 &53.1  \\
$\times$ & \checkmark & \checkmark &$\times$ &41.0 &59.1 &44.7 &23.0 &44.3 &55.7  \\
\checkmark& $\times$ & $\times$ & \checkmark &41.4 &59.6 &45.0 &23.8 &44.4 &55.5  \\
$\times$ & \checkmark & $\times$ & \checkmark & \textbf{41.9}  & \textbf{60.1}  & \textbf{45.8}  & \textbf{24.1}  & \textbf{45.3}  & \textbf{55.8}  \\
\bottomrule
\end{tabular}}
}{
 \caption{Effects of components in exemplar replay. % DS and US denote distribution-preserving selection and our risk-balanced selection, for exemplar set construction. We choose DS as the baseline because it significantly surpasses previous methods. DC and PC represent direct calibration and our partial calibration.
 }
 \label{tab:ablation2}
}
\end{floatrow}
\end{table*}

\begin{table*}[htbp]
\begin{floatrow}
% \setcaptionwidth{2.7in}
\capbtabbox{
\scalebox{0.8}{
\begin{tabular}{c|ccccc}
\toprule
    Setting     & \makecell{DSA}     & $AP$      & $AP_{S}$     & $AP_{M}$     & $AP_{L}$ \\
    \midrule
    \multirow{2}{*}{40+40}
    
    & \textit{w/o}   &39.8  &23.5 &42.8 &53.4  \\
    & \textit{w/} &39.7  &21.6 &42.9 &53.8  \\
    \midrule
    \multirow{2}{*}{70+10}
    
    & \textit{w/o}   &41.7 &24.3 &45.2 &55.7    \\
    & \textit{w/}   &41.9  &24.1 &45.3 &55.8    \\
    \bottomrule
\end{tabular}}
}{
 \caption{Ablation results (\%) for disentangled self-attention (DSA) in the 40+40 and 70+10 settings.}
 \label{tab:mask}
 \small
}

\hspace{0.4in}
% \setcaptionwidth{2.5in}
\capbtabbox{
\scalebox{0.8}{
\begin{tabular}{c|cccc}
\toprule
\makecell{Query No.} & $AP$    & $AP_{S}$     & $AP_{M}$     & $AP_{L}$ \\
\midrule
50&41.5 &23.6 &45.0 &55.7\\
 100&42.0 &24.0 &45.3 &56.6\\
 200&42.2 &24.3 &45.4 &56.6\\
 300&41.9 &24.1 &45.3 &55.8\\
\bottomrule
    \end{tabular}}
}{
 \caption{Ablation results (\%) for the number of expanded queries in the 70+10 setting.}
 \label{tab:query_num}
}

\end{floatrow}
\end{table*}

\noindent\textbf{Effect of disentangled self-attention.} The comparison in Tab.\ref{tab:mask} under both the 40+40 and 70+10 settings shows that removing the self-attention interaction between different query groups has almost no impact on performance. This is reasonable because the class sets detected by different query groups do not overlap, thus eliminating the need for self-attention interaction to remove duplicate predictions. Combined with Fig.\ref{fig:overhead}-right, by disentangling the self-attention computation, the computational complexity can be reduced from quadratic to linear growth without performance drop.

\noindent\textbf{Effect on the number of expanded object queries.} Tab.\ref{tab:query_num} presents the $AP$s with changing number of object queries. It indicates that the expanded queries number has a trivial impact. By expanding fewer queries, we can decrease complexity with negligible harm to performance.

\begin{figure}[tpb]
  \centering
  \includegraphics[height=4.0cm]{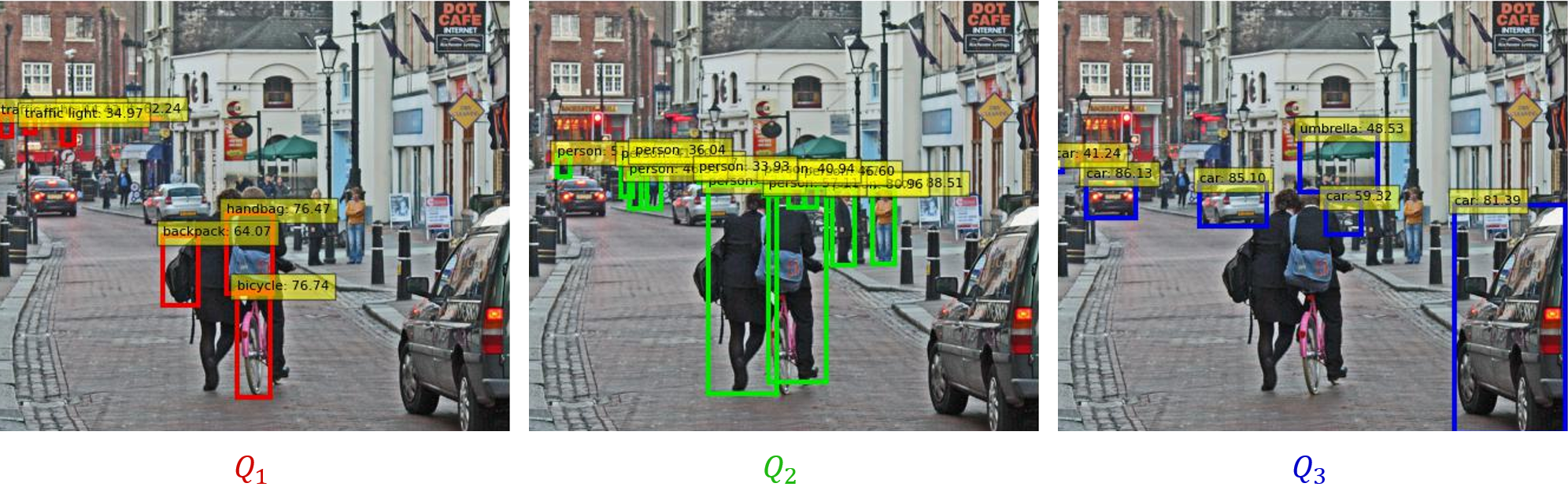}
  \caption{Visualization of the behavior of decoupled queries after training in the 40+20×2 setting. Different query groups are responsible for detecting different sets of classes. More visualization results can be found in the appendix.}
  \label{fig:vis1}
\end{figure}

\noindent\textbf{Visualization of decoupled queries.} In Fig.\ref{fig:vis1}, we visualize the decoupling behavior of  dynamic queries on the test set, based on the 40+20$\times$2 setting. The classes "bicycle", "person", and "car" appear in phases 1, 2, and 3, respectively. They are detected by queries in $Q_1$, $Q_2$ and $Q_3$ accordingly, in a decoupling manner. It can be observed that $Q_1$ detects the "bicycle" class accurately, while $Q_2$ and $Q_3$ consider it as background. Once a query group $Q_t$ learns to detect the class set $C_t$ at step $t$, its class-specific knowledge remains unchanged thereafter, requiring only the maintenance of old knowledge, which significantly improves the performance of IOD.

\section{Conclusion}

In this paper, we propose DyQ-DETR for incremental object detection. Distinct from the mainstream methods focus on the distillation mechanism, we address the catastrophic forgetting by taking inspiration from dynamic networks for model capability expansion. In particular, we propose the dynamic object queries with incremental assembly of new object queries, disentangled self-attention computation and isolated bipartite matching over object queries from different time. The DyQ-DETR can alleviate the conflict between outdated background knowledge and continually emerged classes, thereby achieving stability-plasticity tradeoff. Benefiting from the isolated supervision of dynamic object queries, we further propose risk-balanced partial calibration for effective exemplar replay, with the idea to select exemplars based on risk and partially finetunes the model without relying on low-quality pseudo labels. Extensive experiments demonstrate that our proposed DyQ-DETR surpasses existing IOD methods by a large margin, with quite limited memory overhead. Besides of dynamic query, we hope that more various ways of model expansion can be explored in IOD.

\printbibliography

%%%%%%%%%%%%%%%%%%%%%%%%%%%%%%%%%%%%%%%%%%%%%%%%%%%%%%%%%%%%

\appendix

\section{Appendix}

In Sec.\ref{sec:algorithm}, we present the algorithmic pipeline of DyQ-DETR. In Sec.\ref{sec:ER}, we introduce a simple but strong baseline with only exemplar replay to emphasize the importance and rationale of the non-exemplar setting. In Sec.\ref{sec:more_results}, we provide more detailed experimental results. This includes details of the traditional protocol (Sec.\ref{sec:traditional}), more fine-grained results about old classes and new classes (Sect.\ref{sec:classes}), and more ablation results (Sec. \ref{sec:threshold}). In Sec.\ref{sec:vis}, we show more visualization results, including images with different levels of risk, behavior of decoupled queries, and performance comparison with CL-DETR. Please note that all experiments, except for those in Sec.\ref{sec:traditional}, are based on the revised protocol, using Deformable DETR on COCO 2017, and we will not specify this further.

\section{Algorithmic Pipeline of DyQ-DETR}
\label{sec:algorithm}

\begin{algorithm}[h]
\caption{DyQ-DETR (the $\tau$-th phase)}
\label{algorithm:1}
\SetAlgoLined
\SetKwInput{KwData}{Input}
\SetKwInput{KwResult}{Output}
    \KwData{new class data $D_{\tau}$; old class exemplars $\epsilon_{1:\tau-1}$; old model $\Phi^{old}$.}
    \KwResult{new model $\Phi$; new exemplar set $\epsilon_{1:\tau}$.}
    % Let $argmiddle(y(x),R)$ the function of selecting corresponding x of the R middle y\;
    Let $\Phi \leftarrow \Phi^{old}$\;
    Dynamically expand new queries, i.e., $\{Q^{old}, Q^{new}\} \leftarrow \{Q^{old}\}$\;
    Disentangle self-attention, i.e., $attn(Q_{m,i}, Q_{n,j})=0, m \neq n$\;
    
    \For(\tcp*[f]{dynamic query for incremental training}){\emph{epochs}} 
    {
    \For{\emph{mini-batches} $(x, y)\in D_{\tau}$}{
    
      Let $\hat{y}^{old}\leftarrow\Phi^{old}(x)$ and get $y^{pseudo}$ from $\hat{y}^{old}$ \; 
      Let $\hat{y}\leftarrow\Phi(x)$\;
      Compute the decoupled loss $\mathcal{L}_{total}^{DETR} (\hat{y}, y^{pseudo} \cup y)$\;
      Update $\Phi$ via optimizer\;
      }
    }
    
    \For(\tcp*[f]{score by risk}){\emph{mini-batches} $(x, y)\in D_{\tau}$ }{
      
      Let $\hat{y} \gets \Phi(x)$ and select $\tau$-th part $\hat{y}_{\tau}$ from $\hat{y}$\;
      Compute risk score using $\mathcal{L}_{partial} (\hat{y}_{\tau}, y)$
      \;
      }
    Sort $D_{\tau}$ by risk score, and select the middle 10\% as $\epsilon_{\tau}$ \;
    Let $\epsilon_{1:\tau}\leftarrow{\epsilon_{1:\tau-1}}\cup \epsilon_{\tau}$\;
    
  \For(\tcp*[f]{partial calibration}){\emph{epochs}}{
    \For{\emph{mini-batches} $(x, y)\in\epsilon_{1:\tau}$}{
      Let $\hat{y} \gets \Phi(x)$ and select the corresponding part $\hat{y}_{k}$ from $\hat{y}$\;
      Compute partial calibration loss $\mathcal{L}_{partial} (\hat{y}_{k}, y)$ \;
      Update $\Phi$ via optimizer\;
      }
    }
\end{algorithm}

\begin{table*}[t]
  \centering
  \scalebox{1}{
  \begin{tabular}{c|ccccccc|c|c}
    \toprule
    Setting     & ER     & $AP$  & $AP_{50}$  & $AP_{75}$      & $AP_{S}$     & $AP_{M}$     & $AP_{L}$ & $AP_{old}$ & $AP_{new}$ \\
    \midrule
    \multirow{2}{*}{40+40}
    
    & \textit{w/o}   &20.1 &29.6 &21.8 &11.5 &22.9 &26.4  &0 &40.5\\
    & \textit{w/} &31.4 &46.4 &34.0 &16.5 &34.1 &42.2  &26.1 &37.7\\
    % \midrule
    % \multirow{2}{*}{70+10}
    
    % & \textit{w/o}   & &&  & & &  &&\\
    % & \textit{w/}   & &&  & & &  &&\\
    \bottomrule
\end{tabular}}
  \caption{Results (\%) of the simple baseline in the 40+40 setting.}
  \label{table:baseline}
\end{table*}

To get a more comprehensive understanding of the overall framework, we describe the training pipeline of DyQ-DETR in Algorithm \ref{algorithm:1}. In the $\tau$-th phase, we dynamically add a new set of queries, which are responsible for detecting and recognizing new classes. We disentangle self-attention modules in the decoder, ensuring that queries between different groups remain relatively independent. In the incremental training, we use a score threshold to filter out background predictions, resulting in the pseudo-labels $y^{pseudo}$. Instead of simply merging pseudo-labels $y^{pseudo}$ and incomplete real labels $y$, we divide the completed annotations of an image according to the incremental phase/class set. For an image, annotations corresponding to different class sets act on the outputs of respective queries, thus resulting in the decoupled loss $\mathcal{L}_{total}^{DETR}$ for training.

After the incremental training, we utilize the trained model to calculate the partial loss as the risk score for images in $D_{\tau}$. We sort $D_{\tau}$ based on the risk score and simply select the middle 10\% of these samples to serve as exemplars. Then, for an image in the exemplar set $\epsilon_{1:\tau}$, we identify the specific 
stage corresponding to its annotation, and we use the respective output and real labels to compute the partial loss $\mathcal{L}_{partial}$ for calibration.

\section{Baseline with only Exemplar Replay}
\label{sec:ER}

We introduce a simple baseline, which only combats forgetting by saving exemplars of the same size. As shown in Tab.\ref{table:baseline}, it completely forgets past knowledge after the incremental training, but its performance significantly improves after the simple exemplar replay. The total memory budget for exemplars is typically set as 10\% of the total dataset size. This relatively large proportion may obscure issues in the incremental training and potentially diminish the advantages of superior methods. Therefore, we believe that comparisons in non-exemplar settings are more suitable for evaluating the performance of different methods.

\section{More Detailed Experimental Results}
\label{sec:more_results}

\subsection{Traditional protocol and results}
\label{sec:traditional}

\begin{table*}[b]
  \centering
  \scalebox{0.85}{
  \begin{tabular}{lcccccccccccc}
    \toprule
    \multirow{2}{*}{Methods} &
    \multicolumn{4}{c}{All classes} & \multicolumn{4}{c}{Old classes} & \multicolumn{4}{c}{New classes}  \\
    \cmidrule(r){2-5}
    \cmidrule(r){6-9}
    \cmidrule(r){10-13}
    &  $AP$     & $AP_{S}$     & $AP_{M}$    & $AP_{L}$     &  $AP$     & $AP_{S}$     & $AP_{M}$    & $AP_{L}$     &  $AP$     & $AP_{S}$     & $AP_{M}$    & $AP_{L}$   \\
    \midrule
    % LwF \cite{li2017learning} &24.5 &12.4 &28.2 &35.2 &24.0 &12.3 &27.7 &34.4 &-&-&-&-  \\
    % iCaRL \cite{rebuffi2017icarl} &35.9 &19.1 &39.4 &48.6 &36.8 &20.3 &39.9 &50.0 &-&-&-&-  \\
    CL-DETR \cite{liu2023continual} &37.7 &21.1 &40.7 &50.9 &39.7 &22.9 &40.4 &53.7 &36.3 &19.7 &41.4 &49.1  \\
    DyQ-DETR  &39.6 &22.9 &43.0 &53.2 &41.3 &25.3 &42.7 &55.3 &38.6 &21.0 &43.9 &52.4\\
   
    \bottomrule
  \end{tabular}}
  \caption{Fine-grained results (\%) for old and new classes in the 40 + 40 setting.}
  \label{table:table3}
\end{table*}

The main difference between the traditional protocol\cite{shmelkov2017incremental,li2019rilod,peng2021sid,feng2022overcoming} and the revised protocol\cite{liu2023continual} lies in data partitioning. In the traditional protocol, the model can observe all images containing at least one object of the currently interested classes, which may result in an image appearing across multiple phases. Formally, let $D=\{(x,y)\}$ denotes a dataset with images $x$ and corresponding object annotations $y$. First, we divide the total class set into non-overlapping parts $\{C_1, C_2,..., C_T\}$, one for each incremental phase. For each phase $\tau$, all samples in $D$ retain annotations of $C_{\tau}$ and drop others. The incremental training dataset in phase $\tau$ consists of images that contain at least one annotation for $C_{\tau}$. Additionally, the revised protocol shuffles the class order, while the traditional one does not, leading to different class partitions.

Tab.\ref{main_table} shows that our proposed DyQ-DETR consistently outperforms the state-of-the-art (SOTA) method in the traditional IOD protocol, especially in the non-exemplar settings. Specifically, without exemplars, our DyQ-DETR surpasses the SOTA by 2.2\% AP and 3.7\% AP in the 40+40 and 70+10 settings, respectively. With exemplars, the advantage decreases to 0.4\% AP and 2.0\% AP, respectively. The substantial advantage of our DyQ-DETR in non-exemplar settings demonstrates the effectiveness of dynamic query, which is the core of our method.

\begin{figure}[t]
  \centering
  \includegraphics[height=5cm]{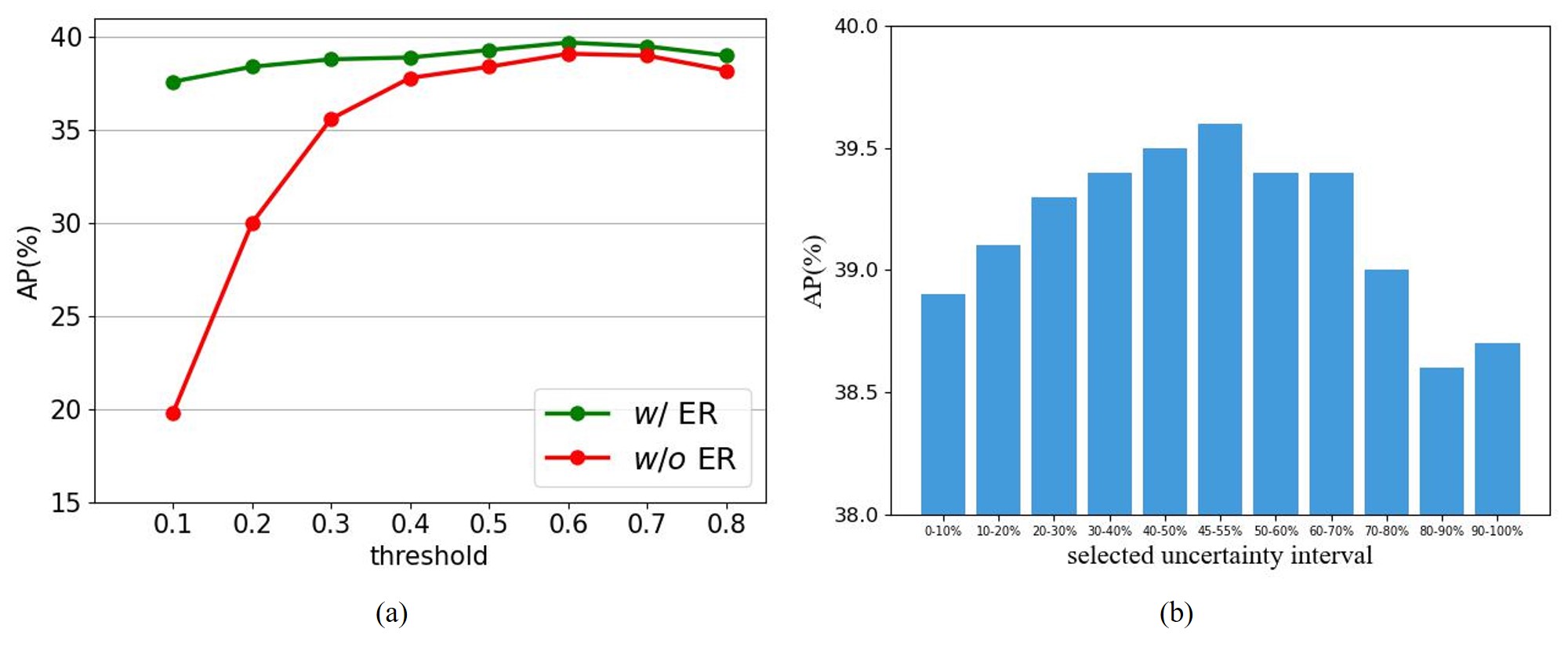}
  \caption{(a) Performance with different score thresholds. (b) Performance with different selected risk intervals. Both experiments are conducted in the 40+40 setting.}
  \label{fig:threshold}
\end{figure}

\subsection{Results about old classes and new classes}
\label{sec:classes}
Tab.\ref{table:table3} presents fine-grained results in the 40+40 setting about old classes and new classes, which respectively represent the model's stability and plasticity. The results show that, compared to the SOTA, our DyQ-DETR achieves improvements of 1.6\% AP for old classes and 2.3\% AP for new classes, indicating enhanced stability and plasticity in our method. 

\subsection{More ablation results}
\label{sec:threshold}

\noindent \textbf{Effect of the score threshold.} In the 40+40 setting, we set the score threshold to 0.6, while for other settings, it is set to 0.4. This is because, in other settings, there is a higher proportion of old classes where stability is preferred, and more old class pseudo-labels help combat forgetting. The ablation study on the score threshold is shown in Fig.\ref{fig:threshold}(a), where the optimal threshold for the 40+40 setting is 0.6. We also find that when the score threshold is low, the pseudo-label noise is significant, leading to poor performance in non-exemplar settings. However, after exemplar replay, the performance greatly improves, further illustrating that exemplar replay may mask issues in the incremental training.

\noindent \textbf{Effect of the selected risk interval.} After sorting the images by risk in ascending order, the results of changing the selection interval are illustrated in Fig.\ref{fig:threshold}(b). It can be observed that selecting images with moderate risk is optimal. This selection strategy is straightforward, and our contribution lies more in demonstrating the feasibility of using the model to optimize the selection.

\section{More visualization results}
\label{sec:vis}

\begin{figure}[hb]
  \centering
  \includegraphics[height=6.3cm]{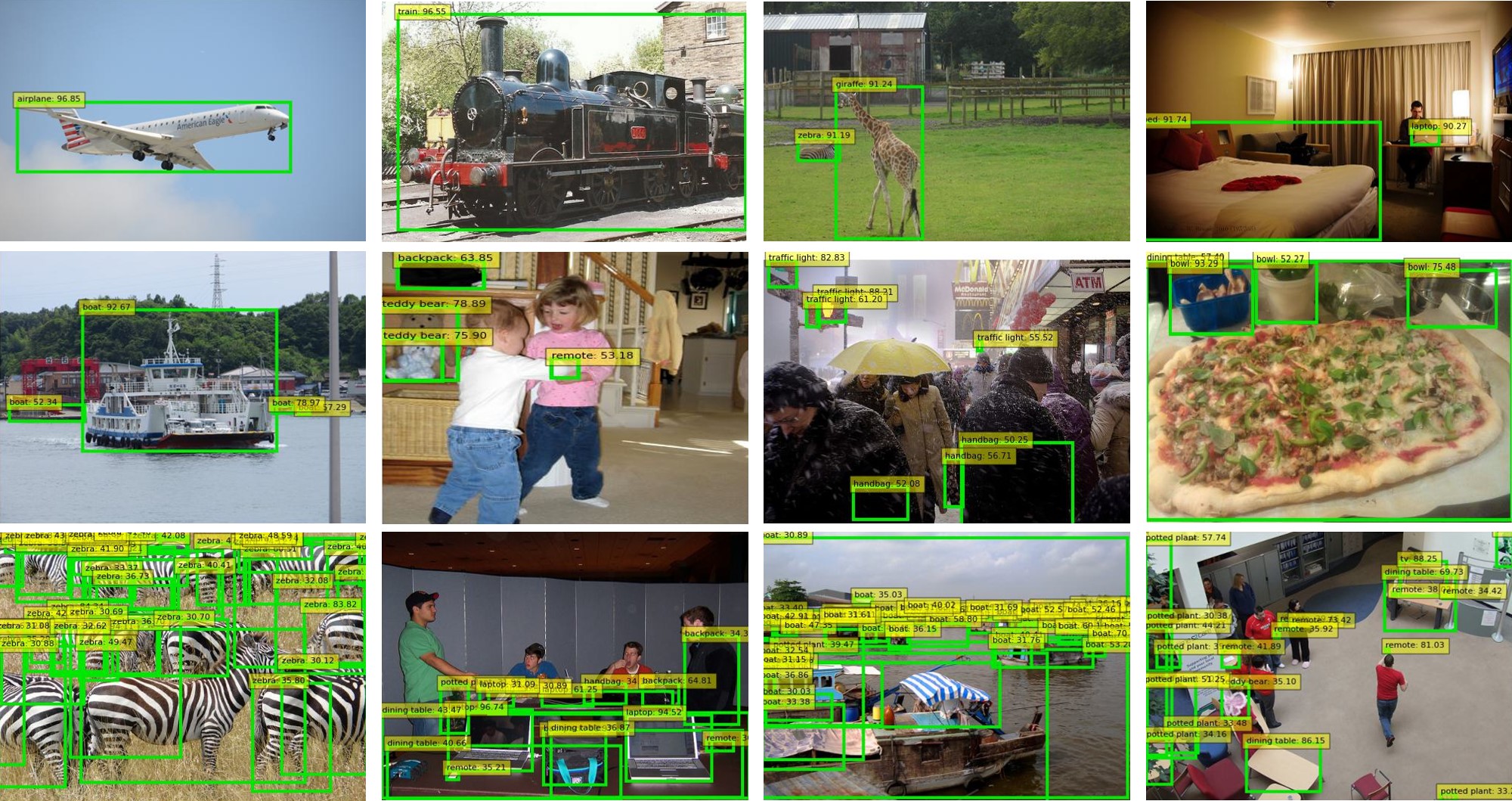}
  \caption{Supplementary to Fig.\ref{fig:risk} (main paper). Visualization of images with different levels of risk. Rows 1, 2, and 3 respectively represent images with low risk, moderate risk (selected by our DyQ-DETR), and high risk.}
  \label{fig:vis2}
\end{figure}

\noindent \textbf{Images with different levels of risk.} As illustrated in Fig.\ref{fig:vis2}, images with low risk are considered as simple samples containing only a few annotated objects, while images with high risk are characterized by a dense distribution of objects, accompanied by severe occlusion or inaccurate annotations. Samples in the middle part are considered informative and reliable, 
and such an active selection strategy promotes more effective exemplar replay.

% Images with low risk appear frequently in the dataset, often containing only one object that occupies the majority of the image, with the object exhibiting representative characteristics of its category. 

% Images with high risk are characterized by dense distribution of objects accompanied by severe occlusion or inaccurate annotations. 

\begin{figure}[hb]
  \centering
  \includegraphics[height=15cm]{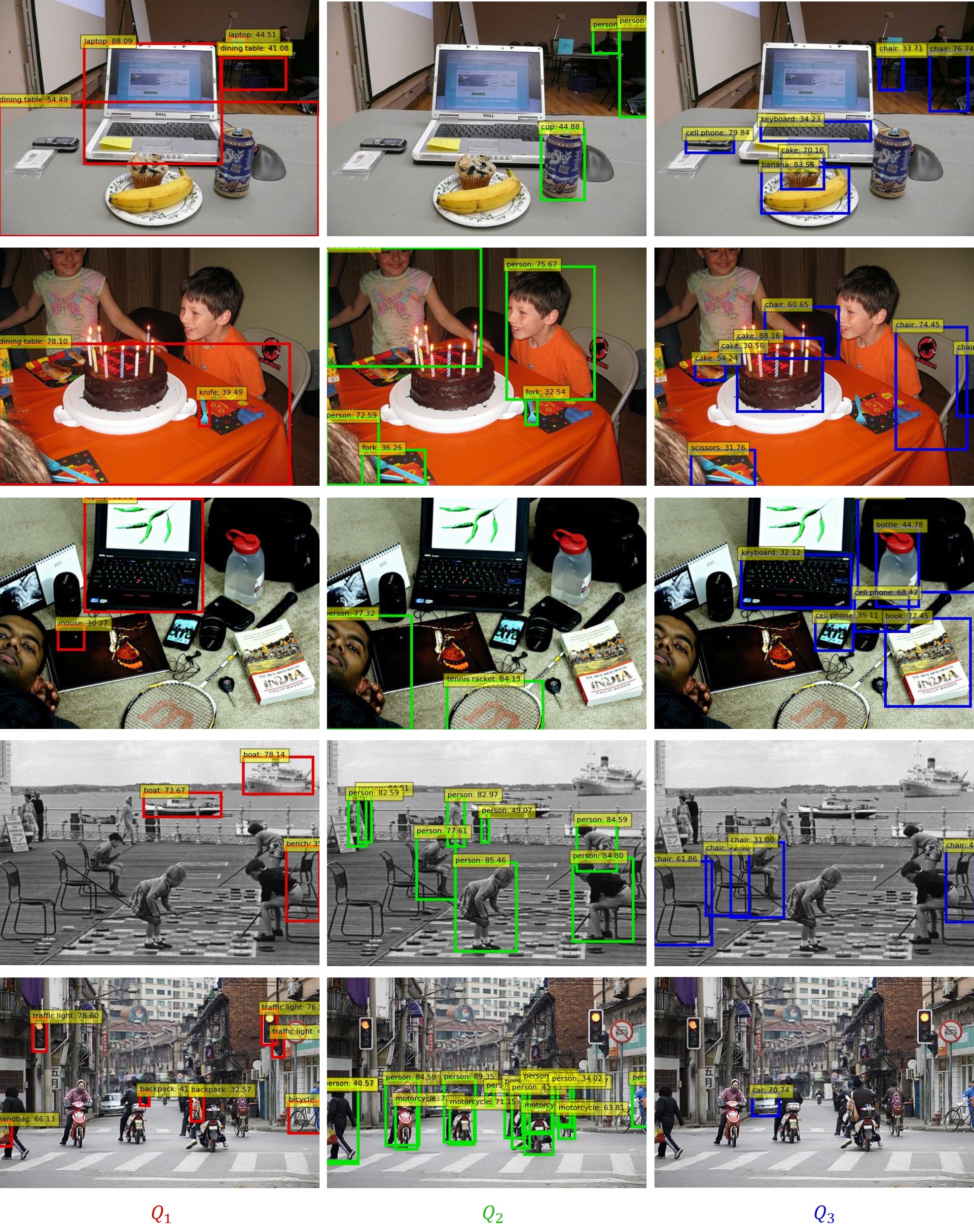}
  \caption{Supplementary to Fig.\ref{fig:vis1} (main paper). Visualization of the behavior of decoupled queries after training in the revised protocol 40+20×2 setting.}
  \label{fig:vis10}
\end{figure}

\noindent \textbf{Behavior of decoupled queries.} As depicted in Fig.\ref{fig:vis10}, our designed dynamic query set achieves the goal of only detecting the corresponding class set. The old query set is responsible for memorizing knowledge of old classes, while the new query set focuses on learning new knowledge. This decoupling significantly improves the performance in non-exemplar settings.

\begin{figure}[t]
  \centering
  \includegraphics[height=12.3cm]{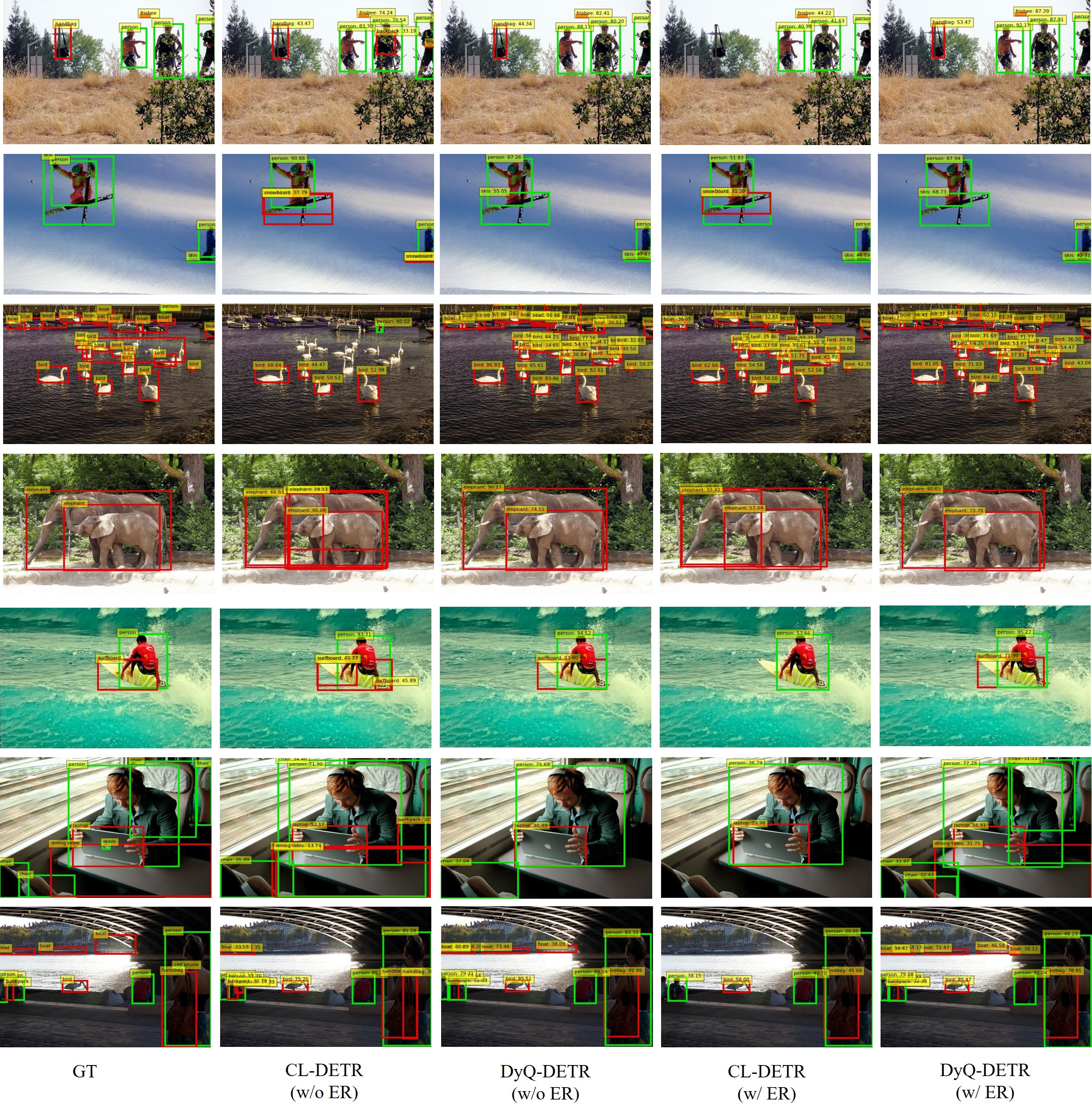}
  \caption{Visualization of performance comparison with CL-DETR, including the non-exemplar setting and exemplar-based setting. Visualization is based on the 40+40 setting. In this figure, red and green represent old classes and new classes, respectively.}
  \label{fig:vis3}
\end{figure}

\noindent \textbf{Performance comparison with CL-DETR.}
Fig.\ref{fig:vis3} shows the visualized detection results of CL-DETR and our DyQ-DETR in non-exemplar and exemplar-based settings. In both settings, DyQ-DETR presents more robust performance for remaining old knowledge and learning new knowledge, thus greatly mitigating catastrophic forgetting.

\section{Discussion over Open Vocabulary Object Detection}
\label{sec:OVD}

Open vocabulary object detection (OVD) leverages visually related language data as auxiliary supervision to bridge the gap between base categories and novel label spaces\cite{wu2024towards,zareian2021open}. By using image-caption pairs, it acquires an extensive vocabulary of concepts, allowing the model to learn object detection with annotations only for certain base categories. Consequently, the model can detect both base classes and novel classes that were not present in the training annotations. OVR-CNN\cite{zareian2021open} is the pioneering work that introduced the concept of open vocabulary object detection and established a comprehensive framework to address this problem. Subsequently, with the introduction of large-scale visual language pretraining models such as CLIP\cite{radford2021learning}, novel approaches have emerged that utilize these pretrained models to enhance the capability of open vocabulary object detection. Some works\cite{gu2021open,pham2023lpovod,rasheed2022bridging,ma2022openvocabulary} employ knowledge distillation, using large model text encoders to align visual features with the large model and transfer new class information. Other works address the gap in object detection region localization by converting image-text level pre-training to region-text level\cite{li2022grounded,zhong2022regionclip,buettner2023investigating,zhang2022glipv2}, while some complete this during the detector training stage after pre-training\cite{VLDet,Zang_2022,yao2023detclipv2,ma2023codet}.

The difference between IOD (Incremental Object Detection) and OVD (Open Vocabulary Detection) lies in their focus. IOD emphasizes the continuous learning and updating process of the model, while OVD focuses on the model's generalization ability. IOD aims to improve the model through multiple learning steps, whereas OVD aims for the model to generalize perfectly to other problems with just training once. Specifically, after training, OVD can detect some objects in a new category space, even if these objects have not been seen in the annotations of base categories, due to the rich semantic concepts obtained during the image-caption alignment process. However, despite the richness of the learned semantic concepts, OVD is difficult to handle newly generated concepts due to the dynamic nature of the environment and requires re-alignment of image-caption pairs to learn these new concepts. In contrast, IOD simplifies this process by allowing incremental learning of new concept based on newly collected and annotated data, thus avoiding the need for retraining the entire model.

IOD and OVD are not mutually exclusive but complementary. They are both important for open-world practical applications. As mentioned above, OVD still shows deficiency in handling the emergence of new concepts without retraining, whereas IOD simplifies this process, saving time and computational resources. However, IOD struggles to resist catastrophic forgetting of old class semantics during continuous model updates. OVD relies on pre-learned rich semantic concepts for novel classes, without consideration of forgetting issue.

\section{Limitations and Broader Impacts}

We propose a dynamic object queries-based detection transformer (DyQ-DETR) to address catastrophic forgetting in incremental object detection. While it scales well to tasks with an incremental step of 20, DyQ-DETR may face challenges in scaling to tasks with longer incremental steps.

High-performance incremental object detection systems have tremendous potential to have a significant impact on various fields and inspire innovative research approaches in robotics, autonomous driving, and beyond. For example, our proposed DyQ-DETR can progressively enhance the recognition and manipulation capabilities of robots, thereby improving efficiency and productivity in industries and healthcare. Furthermore, considering the significant overhead of joint training, DyQ-DETR helps detection models avoid retraining from scratch, providing an effective and efficient approach for incremental updates of large visual models. As for potential widespread impact, our work has not shown any negative social consequences.

% \begin{figure}[t]
%   \centering
%   \includegraphics[height=10cm]{images/comparison.jpg}
%   \caption{Visualization of performance comparison with CL-DETR, including the non-exemplar setting and exemplar-based setting. Visualization is based on the 40+40 setting. In this figure, red and green represent old classes and new classes, respectively.}
%   \label{fig:vis3}
% \end{figure}

%%%%%%%%%%%%%%%%%%%%%%%%%%%%%%%%%%%%%%%%%%%%%%%%%%%%%%%%%%%%

\clearpage

\end{document}